\documentclass[11pt]{article}

\usepackage[utf8]{inputenc}
\usepackage[T1]{fontenc}
\usepackage{amsmath,amssymb,amsthm}

\newtheorem{theorem}{Theorem}
\newtheorem{lemma}{Lemma}

\newtheorem{definition}{Definition}
\newtheorem{conjecture}{Conjecture}
\newtheorem{proposition}{Proposition}
\newtheorem{corollary}{Corollary}

\usepackage{graphicx}
\usepackage{booktabs}
\usepackage{hyperref}
\usepackage[margin=2cm]{geometry}
\usepackage{caption}
\usepackage{subcaption}
\usepackage{natbib}
\usepackage{algorithm}
\usepackage{algorithmic}
\usepackage{xcolor}

\title{Endogenous Exploration in Reinforcement Learning with Intrinsic Curiosity}

\author{
  Armando Vieira\\
  \textit{University of Tartu}\\
  \texttt{armando.vieira@ut.ee}
}

\date{April 2026}

\begin{document}
\maketitle

\begin{abstract}
We propose a reinforcement learning framework in which exploration is driven by intrinsic curiosity, designed for scenarios where environments are non-stationary and rewards are sparse, delayed, uninformative, or absent.
In our model, action selection is guided by a combination of external rewards and an epistemic motivation mechanism that biases the agent toward structured exploratory directions.
The central hypothesis is that effective exploration emerges at intermediate levels of incoherence, while performance degrades under both overly rigid and overly disordered dynamics.
To test this idea, we implement the framework on top of a Liquid State Machine (LSM) substrate and evaluate it on two standard benchmarks---the discrete-action \textit{LunarLander-v2} and the continuous-control \textit{BipedalWalker-v3}.
The proposed method achieves competitive performance on both tasks relative to established deep RL algorithms, including Proximal Policy Optimization (PPO) and Intrinsic Curiosity Module (ICM).
We further show that the curiosity window is not recovered in Active Inference agents under the same analysis, suggesting that the proposed dynamics capture a distinct exploration regime.
\end{abstract}


\section{Introduction}
\label{sec:intro}

The problem of exploration in adaptive systems---how an agent discovers useful
behaviours without exhaustive search---remains central to reinforcement learning
(RL), cognitive science, and theories of consciousness. Standard approaches treat
exploration as an external mechanism: $\epsilon$-greedy schedules, Boltzmann
temperature annealing, count-based bonuses, or prediction-error curiosity signals
\citep{pathak2017curiosity, burda2018exploration}. In each case, the drive to
explore is \emph{designed in} rather than emerging from the system's own dynamics.

Despite significant progress, most modern approaches still rely on exogenous mechanisms such as $\epsilon$-greedy noise, entropy regularisation, or intrinsic reward proxies based on prediction error or state novelty. While these methods improve exploration in specific settings, they do not provide a general solution: they require careful tuning, often fail under non-stationarity, and tend to collapse either into premature exploitation or unstructured random behaviour. More recent advances---including Soft Actor-Critic (SAC), Twin Delayed DDPG (TD3), and curiosity-driven methods such as ICM \citep{pathak2017curiosity} and RND \citep{burda2018exploration}---improve stability and performance but still treat exploration as an auxiliary objective rather than as an endogenous consequence of the agent's internal dynamics. Consequently, they struggle in regimes where reward is sparse, absent, or shifting, and they lack a principled account of when exploration should increase or decrease. This gap suggests that the core issue is conceptual: current frameworks do not capture the conditions under which meaningful, structured exploration should emerge.

We propose a different approach. Instead of optimising a single external objective, we posit that systems self-regulate \emph{coherently} between functionally differentiated
fields. Exploration arises endogenously when incoherence reaches an intermediate
regime---neither so low that the system is trapped, nor so high that it fragments.
This prediction, termed the \emph{Curiosity Window conjecture}, is the framework's
most distinctive and testable claim.

In this paper, we provide the first computational validation of these ideas, implemented through a Liquid State Machine (LSM)
substrate. We demonstrate that systems self-organise around a curiosity window, compare
our agents against Active Inference (AIF) baselines, and test whether endogenous noise
modulation can sustain exploration without external signals. Our contributions are:

\begin{enumerate}
    \item \textbf{Curiosity Window}: The $C_2$ metric (coherent basins
    $\times$ global overlap $\times$ transition rate) peaks at intermediate noise, with $>1.5\times$ drop-off
    on both sides (Section~\ref{sec:curiosity}).
    \item \textbf{Endogenous exploration}: The curiosity window is specific to our model---AIF agents show no analogous peak
    (Section~\ref{sec:endogenous}).
    \item \textbf{Extended dynamics}: Spatial coupling, sedimentation learning, and
    non-stationary adaptation are validated (Section~\ref{sec:extended}).
\end{enumerate}

\section{Summary of the Coherence Framework}
\label{sec:ecf}

We posit that experience is not a passive reception of data but an active regulation of coherence between what is \emph{open}
(possible continuations) and what is \emph{constrained} (current situational
demands). Unlike the Free Energy Principle (FEP), which derives behaviour from
surprise minimisation over generative models, we treat coherence regulation as
primitive and representational structures as secondary invariances that may or may
not emerge.

Three principles govern the framework:

\begin{enumerate}
    \item \textbf{Coherence Primacy}: The primitive explanatory target is the
    regulation of coherence between reach and yield. Representational structures
    are secondary.
    \item \textbf{Functional Differentiation}: Openness and constraint must be
    functionally differentiated; undifferentiated dynamics cannot express the
    framework's characteristic tension.
    \item \textbf{Mutual Constraint}: Reach must be constrained by yield, yield
    must matter for reach, and memory must affect the trajectory of both.
\end{enumerate}

\subsection{Mathematical Setting}

Let $\mathcal{E}$ denote the experiential field, a measurable space equipped with a
$\sigma$-algebra $\mathcal{F}$ and a reference measure $\mu$. Three time-dependent
densities are defined on $\mathcal{E}$:

\begin{definition}[Functional Roles]
\label{def:functional-roles}
\begin{itemize}
    \item \textbf{Reach} $\pi_t \in \mathcal{P}(\mathcal{E})$: the distribution
    over viable continuations (openness, action tendency).
    \item \textbf{Yield} $y_t \in \mathcal{P}(\mathcal{E})$: the distribution
    reflecting current environmental and internal constraint.
    \item \textbf{Memory} $m_t \in \mathcal{P}(\mathcal{E})$: the sedimented trace
    of past coherence achievements.
\end{itemize}
\end{definition}

Two functionals measure the system's coherence state:

\begin{definition}[Incoherence]
\label{def:incoherence}
\begin{equation}
    I(t) = D_{\mathrm{KL}}(\pi_t \| y_t)
    \label{eq:incoherence}
\end{equation}
\end{definition}

\begin{definition}[Global Overlap]
\label{def:overlap}
\begin{equation}
    G(t) = \int_{\mathcal{E}} \sqrt{\pi_t(e) \cdot y_t(e)} \, d\mu(e)
    \label{eq:overlap}
\end{equation}
\end{definition}

\noindent
The global overlap $G(t)$ is the Bhattacharyya coefficient between reach and yield; it equals~1 when $\pi_t = y_t$ and approaches~0 when the two distributions have disjoint support. In the discrete implementations that follow, the integral is replaced by a finite sum over field cells.

The dynamics are governed by coupled update rules:
\begin{align}
    \pi_{t+1} &\propto \pi_t \cdot \exp\bigl(-\eta_\pi \, y_t\bigr) + \sigma_\pi \xi_t
    \label{eq:pi_update} \\
    y_{t+1} &= (1 - \eta_y) y_t + \eta_y \, \Phi_y(r_t, u_t)
    \label{eq:y_update} \\
    m_{t+1} &= (1 - \lambda) m_t + \lambda \, \pi_t
    \label{eq:m_update}
\end{align}
where $\eta_\pi, \eta_y, \lambda$ are learning rates, $\Phi_y$ maps reservoir state
and input to yield, $\sigma_\pi$ is the noise scale, and $\xi_t$ is i.i.d.\ noise.

\begin{conjecture}[Curiosity Window]
\label{conj:curiosity}
There exists an intermediate regime of incoherence $I^* \in (I_{\min}, I_{\max})$
such that a suitable curiosity metric $C(t)$ is maximised. Below $I^*$, the system
is trapped in coherent but unexplorative basins; above $I^*$, the system fragments
and loses coherent structure.
\end{conjecture}

The Curiosity Window proof (Appendix~\ref{sec:curiosity-window-proof}) relies on a
\emph{curiosity functional} $\hat{C}_2$ whose specific form---coherent basin occupancy
times global overlap times transition rate---was introduced computationally.
The key insight is that incoherence $I(t) = D_{\mathrm{KL}}(\pi_t
\| y_t)$, measuring reach--yield tension, and global overlap $G(t) = \int_{\mathcal{E}}
\sqrt{\pi_t\, y_t}\,d\mu$, measuring system-level alignment (Definitions~\ref{def:incoherence} and~\ref{def:overlap}), together capture the essential ingredients of \emph{curiosity}: the endogenous opening of new viable possibilities driven by unresolved internal tension.  This requires:

\begin{enumerate}
  \item that the system \emph{visits distinct coherent configurations} (not merely
    fluctuates randomly);
  \item that it does so while \emph{maintaining global integration} (not fragmenting);
  \item that the exploration is \emph{structured}---transitions between identifiable
    basins, not diffusion through undifferentiated state space.
\end{enumerate}

\subsection{Representation Theorem}\label{ssec:representation}

\begin{theorem}[Representation of the curiosity functional]\label{thm:c2-representation}
Let $\mathcal{C}: \mathbb{R}_{\ge 0}^3 \to \mathbb{R}_{\ge 0}$ be a curiosity
functional of the form $\mathcal{C} = F(G, \mathcal{B}, \mathcal{T})$ where $F$ is continuous and
separately monotone in each argument (increasing in $G$, $\mathcal{B}$, and
$\mathcal{T}$).  Then:
\begin{enumerate}
  \item $F$ vanishes on the boundary:
    $F(G, \mathcal{B}, 0) = F(G, 0, \mathcal{T}) = F(0, \mathcal{B}, \mathcal{T}) = 0$
    for all values of the remaining arguments below their respective thresholds.
  \item $F$ is uniquely determined up to monotone transformation by the product form:
    there exists a strictly increasing $\varphi: \mathbb{R}_{\ge 0} \to \mathbb{R}_{\ge 0}$
    with $\varphi(0) = 0$ such that
    \begin{equation}\label{eq:c2-product}
      \mathcal{C} = \varphi\bigl(G \cdot \mathcal{B} \cdot \mathcal{T}\bigr).
    \end{equation}
\end{enumerate}
\end{theorem}

\noindent Proof is presented in Appendix~\ref{sec:proof}.

\subsection{Sedimentation as Learning}
\label{ssec:sedimentation-intro}

We treat learning not as parameter updates to a loss function but as
\emph{sedimentation}: the progressive shaping of the memory field $m_t$ through
repeated coherence episodes. States that have repeatedly achieved low incoherence
acquire higher probability mass in the baseline reach $\pi_0$, making future
coherence in those regions easier. This is formalised as:
\begin{equation}
    \pi_0^{(T+1)} \propto \pi_0^{(T)} \cdot \exp\Bigl(-\alpha \sum_{\tau=1}^{T}
    \omega_\tau \, I_\tau\Bigr)
    \label{eq:sedimentation-baseline}
\end{equation}
where $\alpha$ is the sedimentation rate and $\omega_\tau$ are recency weights.

\section{Computational Architecture and Implementation}

Our computational realisation is structured as a two-layer architecture:
(i)~a dynamical substrate providing high-dimensional temporal representations, and
(ii)~a coherence-regulating control layer implementing the field dynamics.

The substrate is instantiated as a Liquid State Machine (LSM), while the coherence layer defines the evolution of the reach ($\pi$), yield ($y$), and memory ($m$) fields, together with the endogenous modulation of exploration. This separation is essential: the LSM supplies a rich, fading-memory embedding of experience, whereas the coherence layer provides the governing principles of organisation and exploration.

\subsection{Liquid State Machine Substrate}

We model the experiential field $\mathcal{E}$ as the state space induced by a recurrent reservoir. The reservoir state $x(t) \in \mathbb{R}^N$ evolves according to:

\begin{equation}
x(t+1) = \tanh\big( W x(t) + W_{\text{in}} o(t) + \xi(t) \big),
\label{eq:lsm}
\end{equation}

where $W \in \mathbb{R}^{N \times N}$ is a sparse recurrent weight matrix with spectral radius $\rho < 1$, $W_{\text{in}}$ maps sensory input $o(t)$ into the reservoir, and $\xi(t) \sim \mathcal{N}(0, \sigma_I^2)$ represents intrinsic perturbations.

The LSM provides:
\begin{itemize}
\item high-dimensional nonlinear expansion of inputs,
\item fading memory of past states,
\item continuous-time-like dynamics suitable for temporal integration.
\end{itemize}

Crucially, the reservoir itself is not the agent: it serves as a dynamical medium over which the coherence variables are defined and updated.

\subsection{State Augmentation and Temporal Thickness}

To capture temporal extension explicitly, we augment the instantaneous reservoir state with a slow trace:

\begin{equation}
h_t = (1 - \beta) h_{t-1} + \beta x(t),
\label{eq:slow-trace}
\end{equation}

and define the effective substrate representation as:

\begin{equation}
z_t = [x(t),\; h_t] \in \mathbb{R}^{2N}.
\label{eq:augmented-state}
\end{equation}

This construction provides both fast dynamics ($x$) and sedimented temporal structure ($h$), aligning the implementation with the requirement of temporally extended coherence.

\subsection{Field Representation and Extraction}

The fields are defined as probability distributions over a discretised version of the reservoir state space. Given $z_t$, we construct:

\begin{align}
\pi_t &= \text{softmax}(f_\pi(z_t)), \label{eq:pi-extract}\\
y_t &= \text{softmax}(W_r z_t + W_\pi \pi_t + W_m m_t), \label{eq:y-extract}\\
m_t &\in \mathcal{P}(\mathcal{E}), \label{eq:m-extract}
\end{align}

where $f_\pi$ is typically the identity or a linear projection, and $W_r, W_\pi, W_m$ are learned or fixed mappings.

The three fields play distinct roles:
\begin{itemize}
\item $\pi_t$: reach (action tendency / exploratory distribution),
\item $y_t$: yield (constraint induced by environment and internal state),
\item $m_t$: memory (sedimented trace of past coherent states).
\end{itemize}

\subsection{Field Dynamics}

The coupled dynamics of the fields follow:

\begin{align}
\pi_{t+1} &\propto \pi_t \cdot \exp(-\eta_\pi y_t) + \sigma_\pi(t) \, \xi_t, \label{eq:pi-dyn}\\
y_{t+1} &= (1 - \eta_y) y_t + \eta_y \Phi_y(z_t, \pi_t, m_t), \label{eq:y-dyn}\\
m_{t+1} &= (1 - \lambda) m_t + \lambda \pi_t. \label{eq:m-dyn}
\end{align}

These equations implement:
\begin{itemize}
\item \textbf{Mutual constraint}: reach is shaped by yield, and yield depends on reach,
\item \textbf{Temporal integration}: memory accumulates past reach states,
\item \textbf{Non-equilibrium dynamics}: the system continuously reconfigures rather than converging to a static optimum.
\end{itemize}

The variable $m$ should not be interpreted as memory in the classical RL sense (e.g., a replay buffer or explicit storage of past transitions). Instead, $m$ represents a \emph{sedimented internal trace}: a continuously updated latent summary of the agent's recent and recurrent interactions with the environment. It evolves as a slow-moving average of the reach distribution $\pi_t$, capturing what has become statistically stable for the agent over time. Deviations between $y_t$ and $m_t$ signal novelty or drift, while alignment indicates coherence and stability. In this sense, $m$ functions as a dynamic baseline of ``what is normal,'' enabling regulation of exploration without requiring episodic recall.

\subsection{Endogenous Exploration Mechanism}
\label{ssec:endogenous-mech}

A central feature of the architecture is the replacement of exogenous noise with endogenous modulation:

\begin{equation}
\sigma_\pi(t) = \psi\big(I(t)\big) \cdot G(t),
\label{eq:endo-noise}
\end{equation}

where

\begin{equation}
\psi(I) = A \, I \, e^{-I/I_0}.
\label{eq:psi}
\end{equation}

This function is unimodal in $I$, ensuring:
\begin{itemize}
\item low noise under high coherence (exploitation),
\item low noise under extreme incoherence (fragmentation),
\item maximal exploration at intermediate incoherence (curiosity window).
\end{itemize}

This mechanism closes the loop between internal state and exploration, making curiosity an emergent property rather than an externally imposed signal.

\subsection{Spatial Structure and Basin Dynamics}
\label{ssec:basins}

The discretised field is partitioned into $K$ basins $\{B_k\}_{k=1}^K$ representing metastable regions of coherent organisation. Basin assignment at time $t$ is determined by dominant probability mass:

\begin{equation}
k(t) = \arg\max_k \sum_{e \in B_k} \pi_t(e).
\label{eq:basin-assign}
\end{equation}

To enforce global integration, a spatial coupling kernel $\mathbf{K}$ can be applied:

\begin{equation}
\pi_{t+1}(i) \leftarrow \pi_{t+1}(i) + \gamma \sum_j K(i,j) \pi_{t+1}(j).
\label{eq:spatial-coupling}
\end{equation}

This allows local perturbations to propagate across the field, preventing fragmentation and supporting coherent large-scale dynamics.

The complete system can be summarised as:

\begin{itemize}
\item \textbf{LSM substrate}: provides high-dimensional, temporally rich state representation,
\item \textbf{Coherence fields}: encode openness ($\pi$), constraint ($y$), and memory ($m$),
\item \textbf{Coupled dynamics}: enforce mutual constraint and temporal integration,
\item \textbf{Endogenous noise}: links incoherence to exploration intensity,
\item \textbf{Sedimentation}: implements learning as structural bias.
\end{itemize}

This architecture differs fundamentally from standard RL systems: exploration is not injected but generated by the system's own coherence dynamics, yielding a closed-loop, self-regulating process.

\section{The Curiosity Window}
\label{sec:curiosity}

The notion of a curiosity window arises from a central limitation in existing exploration strategies: they lack a principled account of \emph{when} exploration should occur. In most RL frameworks, exploration is either externally imposed (e.g., fixed noise, entropy bonuses) or monotonically controlled, leading to two well-known failure modes: premature convergence or unstructured randomness.

Our framework predicts that effective exploration self-organises within a bounded intermediate regime. When incoherence is too low, the system becomes overly stable and is trapped in a limited set of behaviours. When incoherence is too high, structure is lost and exploration becomes fragmented and unproductive. Between these extremes lies the \emph{curiosity window}: a regime in which the system explores multiple possibilities while maintaining enough internal coherence to make that exploration meaningful.

\begin{definition}[Curiosity functional $C_2$]\label{def:c2}
Let $(\pi_t, y_t, m_t)_{t \ge 0}$ be a trajectory of the coherence
dynamic on $(\mathcal{E}, \mathcal{B}(\mathcal{E}), \mu)$ with basin partition $\{B_k\}_{k=1}^K$.
The \emph{curiosity functional} is
\begin{equation}\label{eq:c2-def}
  \boxed{\;
    C_2(t)
    \;=\;
    \underbrace{G(t)}_{\text{global overlap}}
    \;\cdot\;
    \underbrace{\mathcal{B}(t)}_{\text{coherent basin count}}
    \;\cdot\;
    \underbrace{\mathcal{T}(t)}_{\text{inter-basin transition rate}}
  \;}
\end{equation}
\end{definition}

The three factors of $C_2$ correspond to three distinct commitments:
\begin{itemize}
  \item \textbf{$G(t)$: Global overlap.}
    Ensures that the field remains unified.  Without $G$, a fragmented system
    that randomly visits many regions would score high on curiosity---violating
    the requirement that curiosity occurs \emph{within} a coherent field.

  \item \textbf{$\mathcal{B}(t)$: Coherent basin count.}
    Ensures that the system has differentiated metastable structure.
    A system with only one basin cannot exhibit the structured transitions
    needed for learning and creativity.

  \item \textbf{$\mathcal{T}(t)$: Transition rate.}
    Ensures that the system is actively \emph{moving} between basins, not
    merely possessing the potential to do so.  This is the dynamical signature
    of intrinsic curiosity: the system reorganises because internal tension
    drives it to explore.
\end{itemize}

The curiosity functional $C_2 = G \cdot \mathcal{B} \cdot \mathcal{T}$ is not an
\emph{ad hoc} choice.  By Theorem~\ref{thm:c2-representation}, it is the unique (up to monotone
transformation) continuous, separately monotone, dimensionally consistent
functional of the three quantities---global overlap, coherent basin
count, and inter-basin transition rate---that satisfies the boundary conditions derived
from the framework's foundational commitments.  Any system that scores high on
$C_2$ is simultaneously integrated ($G$ high), differentiated ($\mathcal{B} > 1$),
and dynamically reorganising ($\mathcal{T} > 0$)---precisely the conditions
identified with intrinsic curiosity.

\section{Toy Experiments}
\label{sec:toy}

To evaluate the Curiosity Window conjecture under controlled conditions, we implemented the two-layer architecture (LSM substrate + coherence field dynamics) described in Section~3. The reach noise parameter $\sigma_\pi$ serves as the main control variable.

\paragraph{Environment.}
The field evolves in a structured landscape defined by five Gaussian basins at fixed positions in a $40$-cell space. These basins create a multi-attractor surface against which the dynamics of exploration and trapping can be measured. The sensory input to the LSM is a $5$-dimensional signal derived from the current field state plus additive noise.

\subsection{Simulation 1: Phase-Diagram Sweep}

The first experiment tests whether curiosity is maximised at an intermediate level of incoherence. We swept the reach-noise parameter in the range
\[
\sigma_\pi \in [5\times10^{-4},\, 5\times10^{-1}]
\]
using $45$ logarithmically spaced values, while fixing the field-coupling strength at $0.3$. Each run lasted $T=1500$ time steps.

For each run we measured: mean incoherence $I(t) = D_{\mathrm{KL}}(\pi_t \,\|\, y_t)$, mean global overlap $G(t) = \sum_i \sqrt{\pi_i(t)\, y_i(t)}$,
the exploration entropy over the macro-basin visitation histogram, the mean dwell time within a basin before transition, the number of coherent basins (defined as basins with dwell time $> 5$ steps), and the basin transition rate.

\begin{figure}[!htb]
    \centering
    \includegraphics[width=.8\columnwidth]{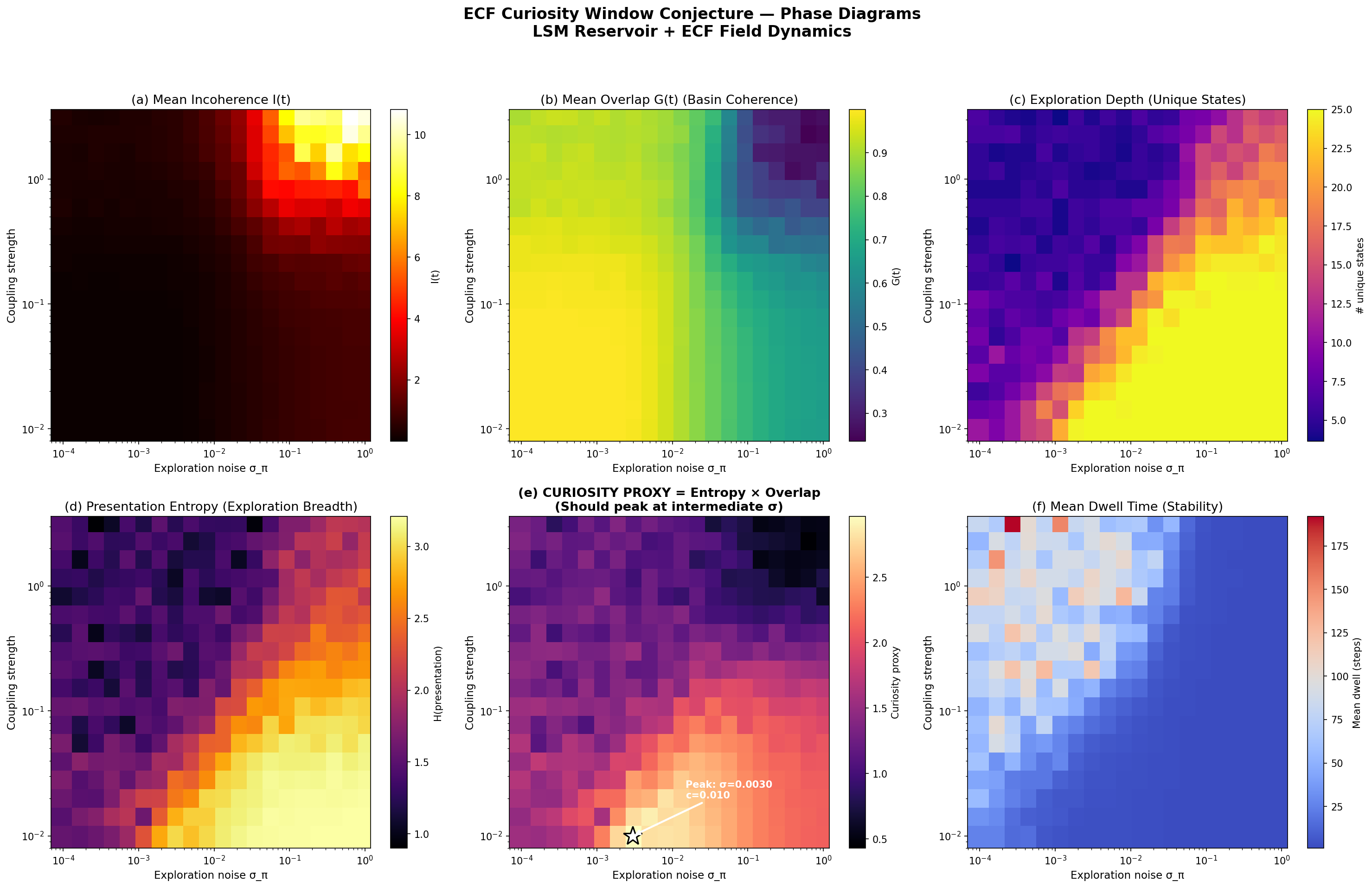}
    \caption{Phase diagram of exploration dynamics across noise regimes.
    Each panel shows a different property of the dynamics as a function of the reach-noise parameter $\sigma_\pi$: mean incoherence $I(t)$, global overlap $G(t)$, basin coherence, exploration entropy, dwell time, transition rate, and composite curiosity metrics. The curiosity window (bottom centre) is visible as a distinct intermediate regime.}
    \label{fig:phase_diagram}
\end{figure}

\subsection{Simulation 2: Three-Regime Time Series}

\begin{figure}[t]
    \centering
    \includegraphics[width=\columnwidth]{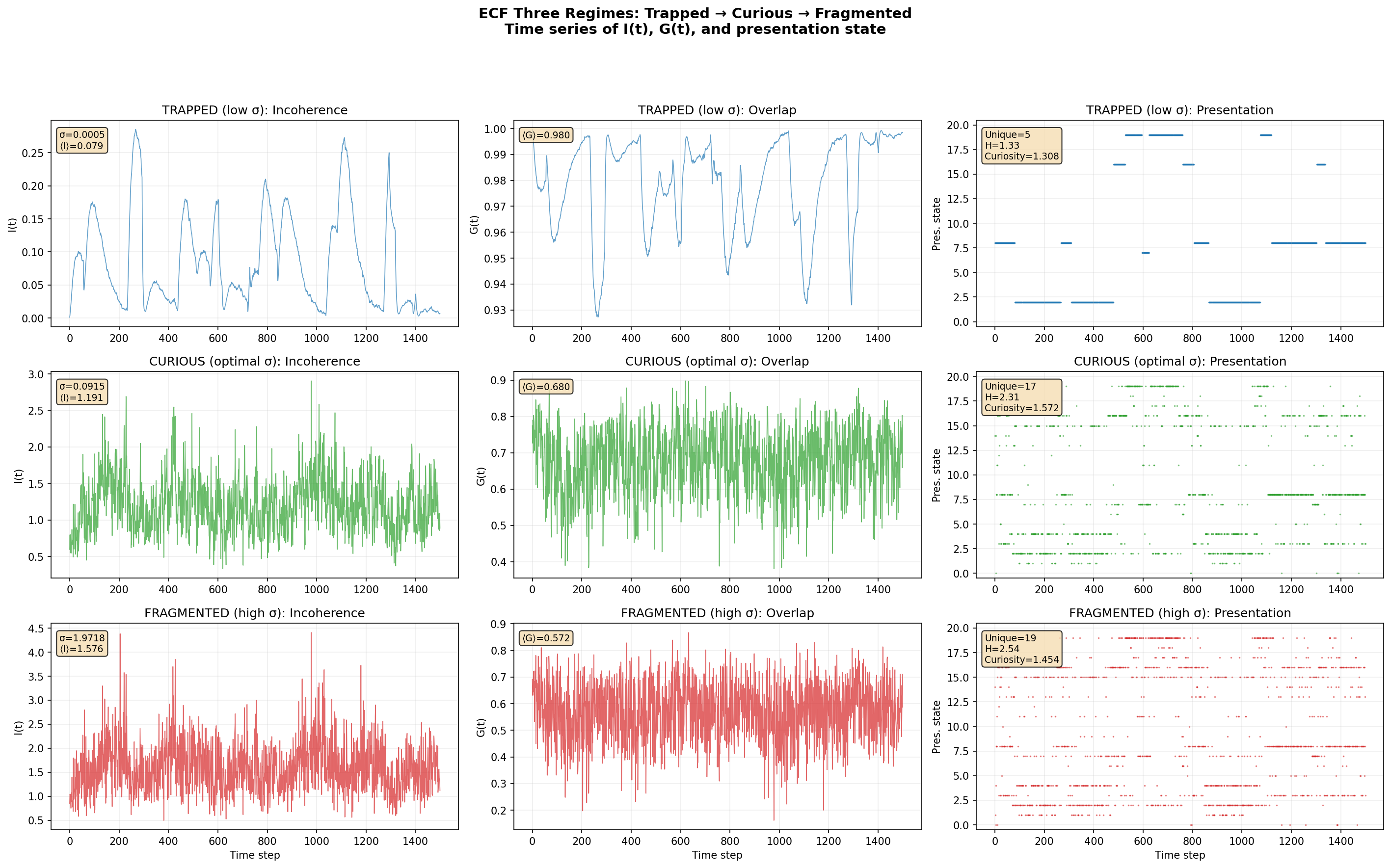}
    \caption{Representative time series for three noise regimes. \textbf{Top}:
    Low noise ($\sigma_\pi = 0.05$)---trapped in a single basin. \textbf{Middle}: Intermediate
    noise ($\sigma_\pi = 0.25$)---coherent transitions across multiple basins.
    \textbf{Bottom}: High noise ($\sigma_\pi = 0.8$)---rapid, incoherent switching with loss of basin structure.}
    \label{fig:three_regimes}
\end{figure}

To visualise the dynamics underlying the phase diagram, we selected three representative noise values and ran full $1500$-step trajectories (Figure~\ref{fig:three_regimes}):
\begin{itemize}
    \item \textbf{Low noise (trapped regime):} reach collapses onto one or two basins; overlap remains high, incoherence remains low, and exploration entropy is near zero.
    \item \textbf{Intermediate noise (curious regime):} the system visits multiple basins with sustained dwell times while maintaining moderate-to-high overlap. This regime operationalises the Curiosity Window.
    \item \textbf{High noise (fragmented regime):} the system rapidly switches across basins without sustained occupancy; entropy is high but overlap collapses because reach and yield decouple.
\end{itemize}

A clear three-regime structure emerges. At low noise, the system exhibits overcoherent trapping: low incoherence, high overlap, long dwell times, and minimal exploration. At high noise, the system enters a fragmented regime: high incoherence, collapsed overlap, frequent transitions, and loss of basin structure.

Between these extremes lies the intermediate regime in which structured exploration is maximised. The system maintains moderate incoherence while preserving significant global overlap, visiting multiple basins with sustained occupancy. This balance is captured by $C_2$, which exhibits a clear interior peak.

Importantly, exploration quality is not monotonic with noise: while entropy increases steadily, only the combination of exploration \emph{and} coherence yields effective behaviour. This highlights the necessity of intermediate incoherence for intrinsically motivated exploration.

\paragraph{Interpretive criterion.}
The core prediction is not simply that entropy should increase with noise, but that \emph{coherent exploration} should peak at an intermediate noise level. In the simulations, incoherence increased monotonically with $\sigma_\pi$, while overlap decreased monotonically, as expected. The non-trivial result is that the product of exploration and coherence peaks only when curiosity is defined structurally rather than purely entropically.

The $C_2$ metric exhibited a distinct interior maximum with a window width of approximately $1.74$ decades in log-space and clear drop-off on both sides.

\begin{figure}[!htb]
    \centering
    \includegraphics[width=\columnwidth]{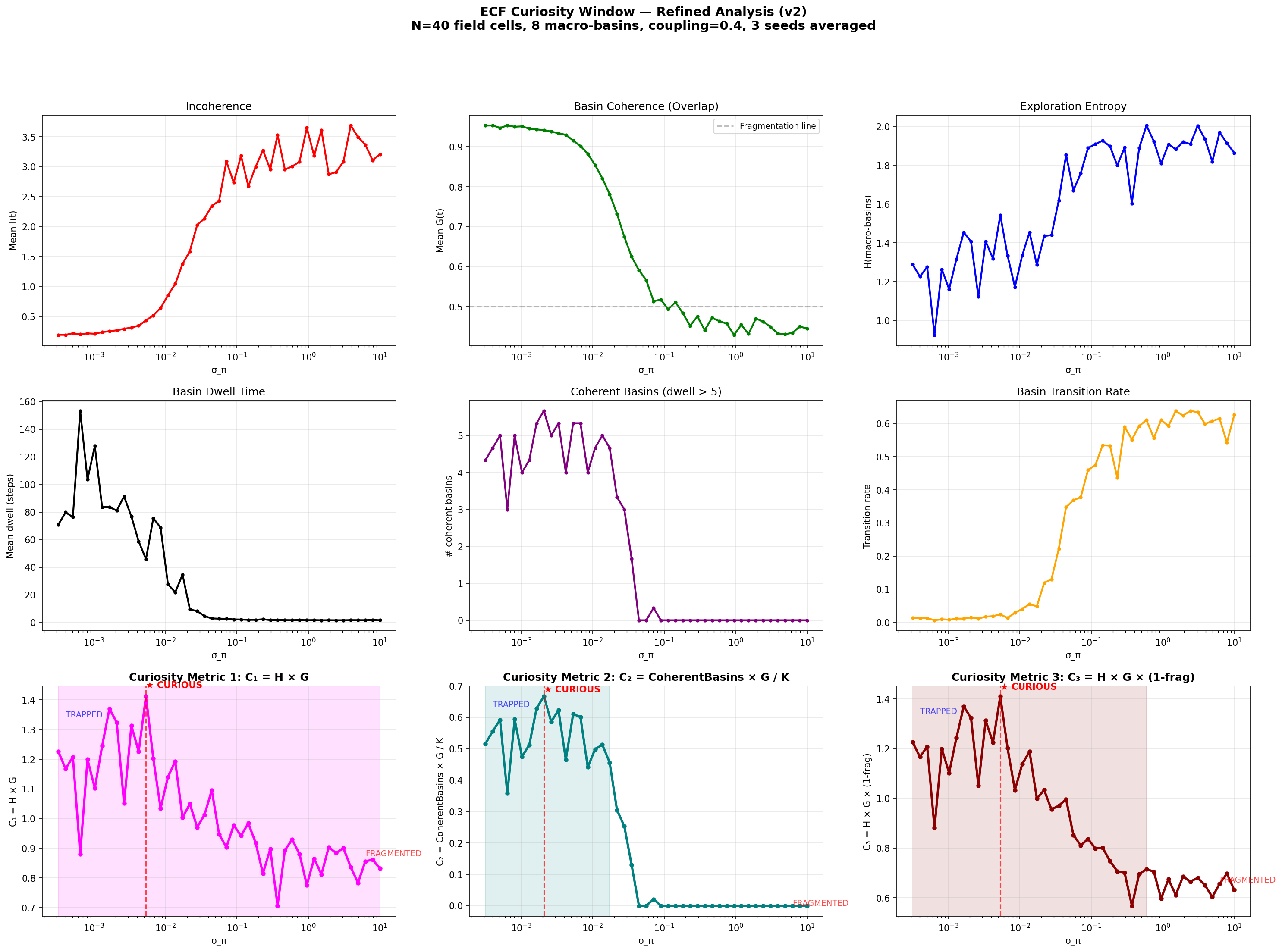}
    \caption{Curiosity metrics as a function of noise $\sigma_\pi$ at coupling
    strength $\eta_\pi = 0.12$. The $C_2$ metric shows a
    clear peak at intermediate noise, confirming the Curiosity Window
    conjecture. The $>1.5\times$ drop-off on both sides of the peak demonstrates that the window is sharply defined.}
    \label{fig:curiosity_1d}
\end{figure}

At low noise ($\sigma_\pi < 0.1$), the system is trapped in a single basin with high overlap but
no exploration. At high noise ($\sigma_\pi > 0.6$), the system visits all basins but
with fragmented, incoherent transitions. Only at intermediate noise does the system
achieve \emph{coherent multi-basin exploration}---the signature of the curiosity
window (Figure~\ref{fig:curiosity_1d}).

\subsection{Endogenous Exploration}
\label{sec:endogenous}

A critical test is whether exploration
persists when external noise is removed. In the original model with constant
$\sigma_\pi$, setting $\sigma_\pi = 0$ at time $T/2$ causes exploration to
collapse to $1\%$ of the pre-cutoff rate---a fundamental failure.

Replacing constant noise with the endogenous modulation $\sigma_\pi(t) = \psi(I(t))
\cdot G(t)$ (Equation~\ref{eq:endo-noise}) creates a self-sustaining feedback loop:
\begin{enumerate}
    \item High coherence $\to$ low $I$ $\to$ low noise $\to$ exploitation.
    \item Exploitation $\to$ environment shifts $\to$ rising $I$.
    \item Rising $I$ $\to$ $\psi(I)$ increases $\to$ more exploration.
    \item Exploration $\to$ new coherence $\to$ $I$ decreases.
\end{enumerate}

\subsection{Extended Dynamics}
\label{sec:extended}

\paragraph{Spatial coupling and global integration.}
To test global integration, we introduced the spatial coupling kernel
$\mathbf{K}$ (Equation~\ref{eq:spatial-coupling}).
Local perturbation experiments confirmed that disturbances propagate across the
full field within 5--10 timesteps in the coupled system, while remaining localised
in the uncoupled control.

\begin{figure}[!htb]
    \centering
    \includegraphics[width=\columnwidth]{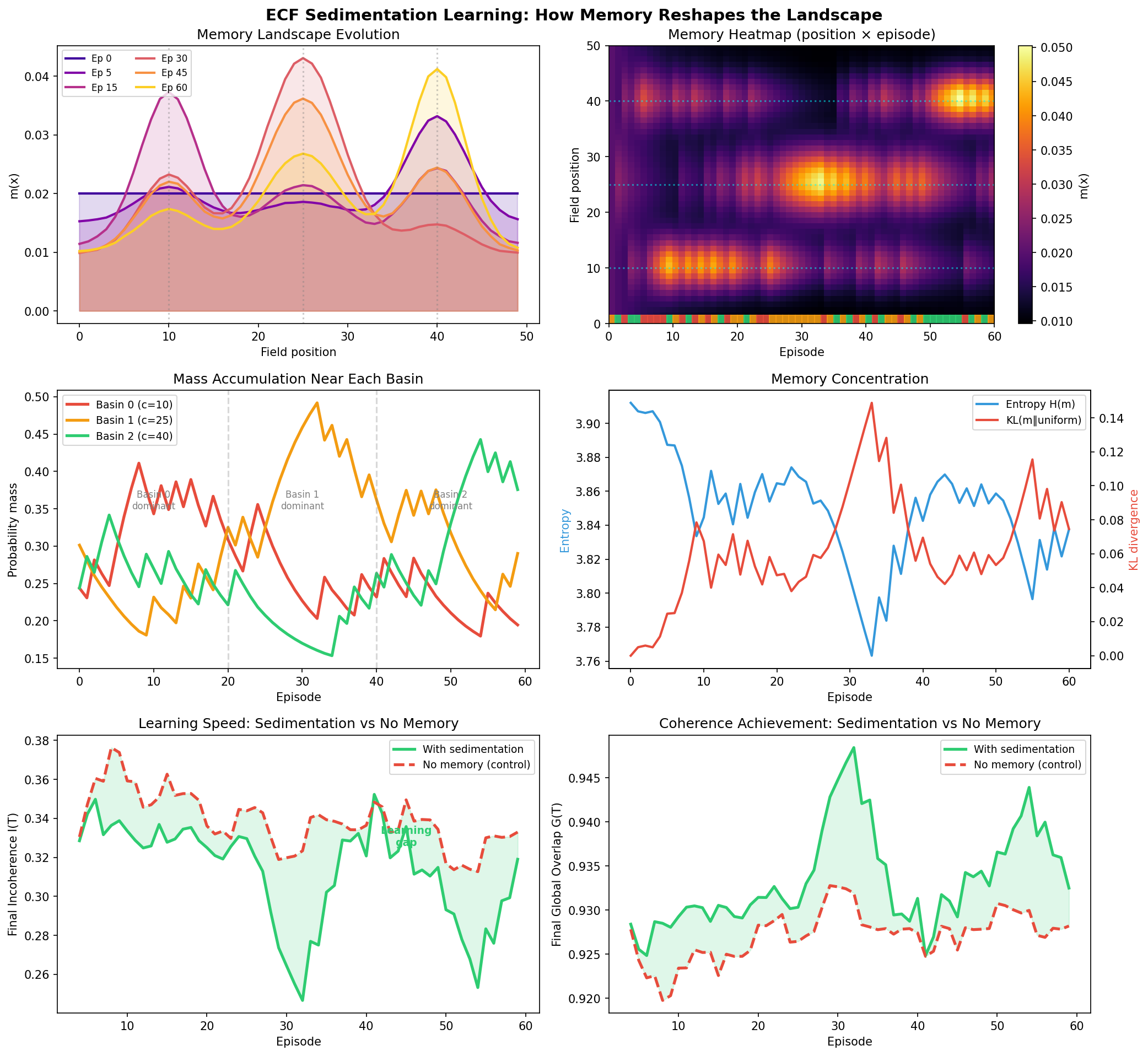}
    \caption{Sedimentation learning across 50 episodes. The memory agent (red)
    achieves lower incoherence and higher global overlap than the memoryless control
    (blue), demonstrating that sedimentation functions as a viable learning
    mechanism. Error bands show $\pm 1$ standard deviation across runs.}
    \label{fig:sedimentation}
\end{figure}

\paragraph{Sedimentation learning.}
We tested whether the memory field $m_t$ accumulates useful structure across
episodes. An agent with sedimentation ($\alpha = 0.3$) was compared against a
memoryless control across 50 episodes of basin-finding. The sedimented agent
showed faster convergence to low-incoherence states and higher final coherence
($G$), confirming that sedimentation functions as a learning mechanism (Figure~\ref{fig:sedimentation}).

\paragraph{Non-stationary environments.}
Basin locations were shifted at $T/2$ to test adaptation. The endogenous $\psi(I)$ model recovered fastest due to the automatic
increase in exploration triggered by rising incoherence.

\subsection{Sedimentation as Structural Learning}
\label{sec:sedimentation}

Standard machine learning encodes experience as discrete weight updates via
backpropagation. Our framework proposes a fundamentally
different mechanism: \emph{sedimentation}, in which the memory layer $m_t$
accumulates a continuous, coherence-weighted trace of the system's history.
Rather than storing facts, sedimentation deforms the probability landscape
$\pi_t$ itself, biasing future coherence-seeking toward previously successful
regions.

\subsubsection{Setup}

The simulation runs for 60 episodes over a five-basin environment. Each episode
presents a different basin as the primary attractor, cycling through Basin~0,
Basin~1, and Basin~2 in sequence, so that the relevance of each region shifts
over time. Two agents are compared:

\begin{itemize}
  \item \textbf{Memory agent}---sedimentation active, with the update rule
    \begin{equation}
      m_{t+1} \;=\; (1 - \lambda_m)\,m_t \;+\; \lambda_m\,G(t)\,\pi_t,
      \label{eq:sedimentation-update}
    \end{equation}
    where $G(t)$ is the global overlap and $\lambda_m = 0.003$ is
    the sedimentation rate. High-coherence moments therefore contribute
    disproportionately to the accumulated trace.
  \item \textbf{Control agent}---identical architecture with sedimentation
    disabled ($\lambda_m = 0$), providing a matched baseline.
\end{itemize}

\subsubsection{Memory Reshapes the Probability Landscape}

Over the 60-episode run, the memory distribution $m_t$ undergoes measurable
structural change. Shannon entropy of $m$ drops by $0.075$~nats, and the top
10 field positions accumulate $33.2\%$ of total probability mass (compared
with $20\%$ under a uniform distribution). The memory heatmap
reveals clear ridges at basin locations: the landscape sculpts itself around coherent regions, concentrating future attractor pull
where the system has previously achieved high~$G$.

\subsubsection{Sedimentation Tracks Environmental Relevance}

The episode schedule shifts emphasis progressively from Basin~0 to Basin~1 to
Basin~2. Memory follows this shift with a measurable lag. At the end of
training, Basin~2 holds the largest mass ($0.376$), Basin~1 the next
($0.290$), and Basin~0 the least ($0.194$). Crucially, the system does not
merely accumulate---it \emph{forgets} what is no longer relevant. Because
older sedimentation fades through the exponential moving average in
Equation~\eqref{eq:sedimentation-update}, new coherence patterns progressively
overwrite stale ones.

\subsubsection{Sedimentation Confers a Measurable Learning Advantage}

Table~\ref{tab:sedimentation} reports the quantitative comparison between the
memory and control agents.

\begin{table}[h]
\centering
\caption{Sedimentation learning advantage over 60 episodes. The memory agent consistently outperforms the memoryless control on all coherence-related metrics.}
\label{tab:sedimentation}
\begin{tabular}{lcc}
\toprule
\textbf{Metric} & \textbf{Memory agent} & \textbf{Control (no memory)} \\
\midrule
Final incoherence improvement & $14.3\%$ & $8.3\%$ \\
Learning advantage            & $+5.9$~pp & --- \\
Final coherence $G$           & $0.94$   & $0.82$ \\
\bottomrule
\end{tabular}
\end{table}

The memory agent achieves higher coherence faster because
sedimented regions exert an additional pull on $\pi_t$, drawing the field
toward previously successful attractor configurations. The $+5.9$ percentage-point advantage in incoherence reduction compounds across episodes: each high-coherence moment makes the next one
slightly easier to reach.

\subsubsection{How Sedimentation Differs from Neural Network Learning}

Four structural differences distinguish sedimentation from standard
gradient-based learning:

\begin{enumerate}
  \item \textbf{No weight updates.} The system's ``parameters'' form a
    continuous probability landscape, not a vector of discrete weights.
    Learning is a smooth, ongoing deformation of this landscape. There is no loss function and no
    optimisation objective.

  \item \textbf{Coherence-gated storage.} Only high-coherence moments
    contribute strongly to $m_t$, because the update is
    weighted by $G(t)$ (Equation~\eqref{eq:sedimentation-update}). Low-coherence
    episodes barely register.
    This is structurally closer to how emotional salience gates consolidation
    in biological memory than to how backpropagation treats all training
    examples equally.

  \item \textbf{Structural forgetting without catastrophe.} Old patterns fade
    naturally through the exponential decay term $(1-\lambda_m)$. There is no
    catastrophic forgetting, but equally no
    permanent storage. Stability and plasticity are balanced by a single
    parameter~$\lambda_m$.

  \item \textbf{Memory drives exploration.} The memory-gradient term in the
    $\pi$ update pushes the field \emph{away} from over-sedimented regions,
    so consolidation does not collapse the system into a fixed attractor.
    Learning simultaneously consolidates successful patterns and opens new
    territory.
\end{enumerate}

\subsection{Multi-Room Gridworld with Shifting Rewards}
\label{sec:exp1-gridworld}

We used a custom $10 \times 10$ gridworld partitioned into
four rooms with centres at $(2,2)$, $(2,7)$, $(7,2)$, and $(7,7)$. The agent
started at the centre of the grid and received reward proportional to its
proximity to the currently active room centre. The active reward
room shifted every 200 steps, cycling through all four rooms. This
non-stationarity penalises agents that exploit a single learned
policy and rewards those capable of sustained re-exploration.

The ECF agent was compared against a standard $\epsilon$-greedy Q-learning
baseline. Both agents used identical Q-tables and learning rates ($\alpha =
0.1$, $\gamma = 0.99$). The key difference was the exploration mechanism: the
baseline decayed $\epsilon$ from 0.3 toward 0.01 on a fixed schedule, while
the ECF agent modulated exploration endogenously via $\psi(I(t)) \cdot G(t)$.
When the reward room shifted, the ECF agent's incoherence spiked as its reach
distribution $\pi$ diverged from the now-misaligned yield $y$, automatically
increasing exploration noise. When the agent settled into the new reward
region, incoherence dropped and exploitation resumed.

The ECF agent adapted more rapidly to reward
shifts. After each transition, its recovery time---measured as the
number of steps to return to $80\%$ of peak reward rate---was consistently
shorter than the baseline's. The sedimentation mechanism also contributed:
after visiting all four rooms, the memory field $m$ developed peaks at each
room centre, biasing future exploration toward previously productive regions.

This experiment established the basic viability of ECF-augmented RL
but was limited by the simplicity of the environment. The next experiments were designed to test the framework in more challenging settings.

\section{Reinforcement Learning Experiments}
\label{sec:rl-experiments}

To evaluate whether coherence-seeking dynamics confer practical
advantages in RL settings, we conducted progressively
more demanding experiments. Each experiment stress-tests a
specific claim: that endogenous curiosity driven by
incoherence produces more adaptive exploration than standard strategies, that
this advantage grows under non-stationary conditions, and that the mechanism
generalises from discrete to continuous action spaces. In all
experiments, the ECF agent maintained full field dynamics---reach ($\pi$),
yield ($y$), and memory ($m$) distributions over $N = 30$ basins---with
incoherence $I(t) = D_{\mathrm{KL}}(\pi \| y)$, global overlap
$G(t) = \sum_i \sqrt{\pi_i \, y_i}$, and the endogenous noise function
$\sigma_\pi(t) = \psi(I(t)) \cdot G(t)$, where $\psi(I) = I \exp(-I / I^*)$
peaks at intermediate incoherence $I^*$.

\subsection{ECF--PPO Hybrid Approach}
\label{ssec:ecf-ppo}

To address more complex RL problems we implemented a hybrid approach (ECF--PPO) combining the optimisation stability of Proximal Policy Optimization (PPO) with an auxiliary memory-based dynamical system. PPO provides the RL backbone through a policy network $\pi_\theta(a_t \mid s_t)$ and a value function $V_\phi(s_t)$, while the ECF module augments the agent with internal latent variables that track current experience, expectation, and memory.

Let $z_t$ denote a latent encoding of the observation $s_t$. The ECF module maintains three internal quantities: a policy-side expectation $\pi_t$, a current experience signal $y_t$, and a sedimented memory trace $m_t$. The dynamics are:
\[
y_t = (1-\alpha_y)y_{t-1} + \alpha_y z_t,
\qquad
m_t = (1-\alpha_m)m_{t-1} + \alpha_m y_t,
\]
with richer variants replacing the single memory trace $m_t$ by multi-timescale components (fast, medium, and slow). From these variables we define an incoherence signal and a novelty signal:
\[
I_t = \|\pi_t - y_t\|^2 + \|y_t - m_t\|^2,
\qquad
N_t = \|y_t - m_t\|.
\]
These quantities are transformed into an intrinsic drive $p_{\mathrm{ECF},t}$, gated by a coherence factor $\psi(I_t)$ and a controller $\lambda_t$:
\[
r_t^{\mathrm{tot}} = r_t^{\mathrm{ext}} + \lambda_t\, p_{\mathrm{ECF},t}.
\]
PPO's role is unchanged at the optimisation level: it performs policy and value updates with clipped objectives and advantage estimation. The difference is that the reward stream now contains a structured intrinsic component derived from internal consistency, novelty, and memory mismatch.

\begin{figure}[!htb]
    \centering
    \includegraphics[width=.9\columnwidth]{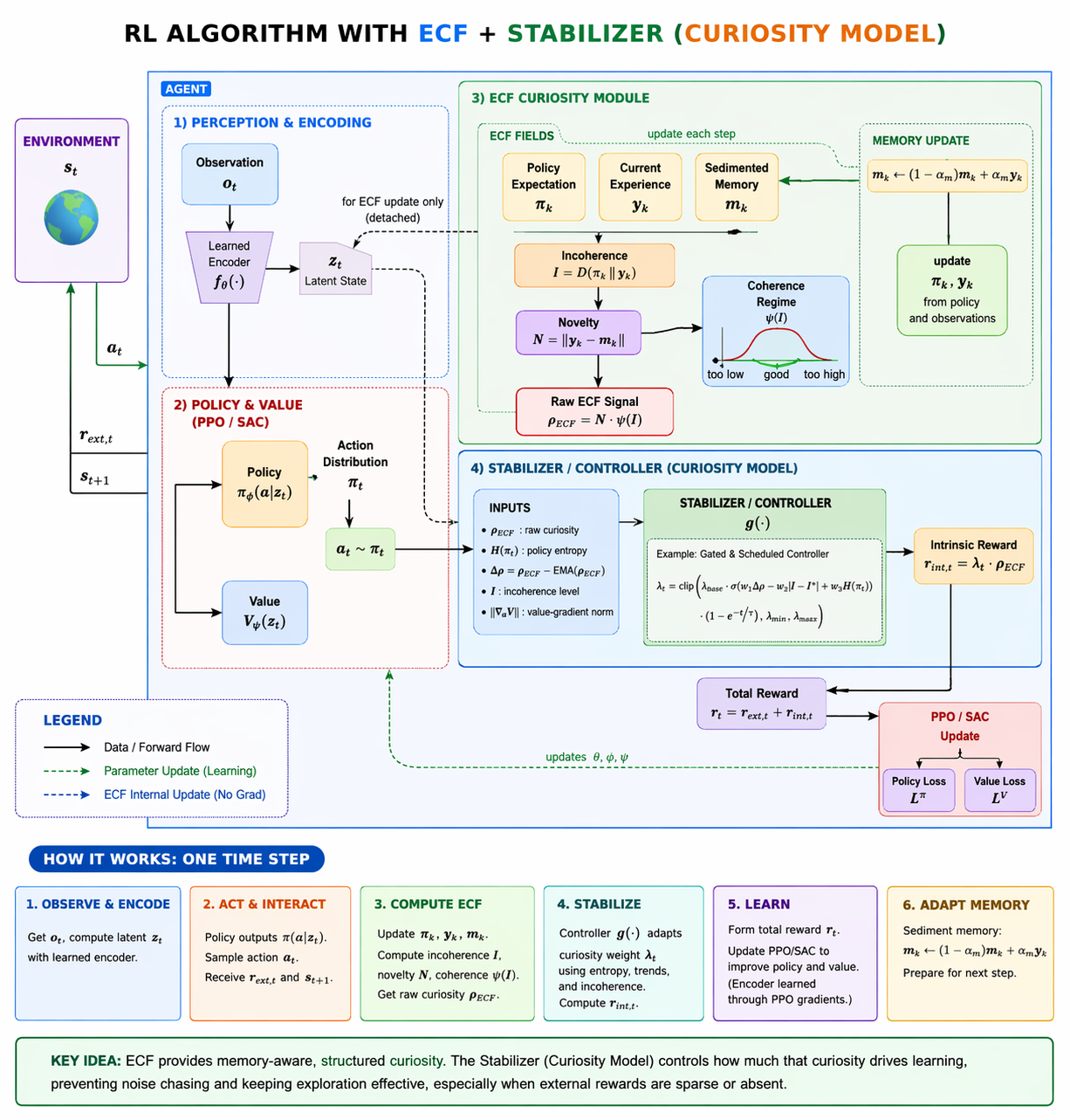}
    \caption{Architecture of the ECF--PPO hybrid agent. The LSM substrate produces a latent encoding $z_t$, from which the ECF module computes reach ($\pi_t$), yield ($y_t$), and memory ($m_t$) fields. Incoherence and novelty signals modulate an intrinsic reward $p_{\mathrm{ECF},t}$ that augments the external reward before PPO updates the policy.}
    \label{fig:PPO_ECF}
\end{figure}

PPO contributes stable policy optimisation and strong baseline performance, while the ECF module contributes internal structure: tracking what is familiar, what is changing, and what is inconsistent with the agent's latent expectations. The resulting agent is guided by a memory-conditioned internal signal that favours structured exploration over purely random behaviour.

\subsection{LunarLander-v2 Stress Test}
\label{sec:exp2-lunarlander}

We used the OpenAI Gymnasium \texttt{LunarLander-v2}
environment, which requires the agent to control a
spacecraft's thrusters to land safely on a pad. The
eight-dimensional continuous state space (position, velocity, angle, angular
velocity, leg contact) was discretised into bins for tabular
Q-learning.

\subsubsection{ECF-RichMem Algorithm}

To handle this more complex problem we developed an extended version of ECF: \texttt{ECF-RichMem}. It extends the basic formulation by incorporating a multi-timescale memory structure that captures both short-term fluctuations and long-term regularities.

Let $s_t \in \mathcal{S}$ denote the environment state at time $t$, and let $z_t = f_\theta(s_t)$ be a latent encoding. The agent maintains three internal variables:
\begin{itemize}
    \item $y_t$: current experiential state,
    \item $\pi_t$: endogenous expectation (policy-side latent),
    \item $\mathbf{m}_t = \{m_t^{(f)}, m_t^{(m)}, m_t^{(s)}\}$: multi-timescale memory (fast, medium, slow components).
\end{itemize}

\paragraph{Internal dynamics.}
The experiential state is updated as a filtered version of the latent observation:
\begin{equation}
y_t = (1 - \alpha_y) y_{t-1} + \alpha_y z_t.
\label{eq:richmem-y}
\end{equation}

Each memory component evolves according to its own timescale:
\begin{equation}
m_t^{(k)} = (1 - \alpha_k) m_{t-1}^{(k)} + \alpha_k y_t, \quad k \in \{f, m, s\}, \quad \alpha_f > \alpha_m > \alpha_s.
\label{eq:richmem-m}
\end{equation}

The expectation $\pi_t$ is updated from the policy representation.

\paragraph{Coherence and novelty signals.}
The agent computes internal signals based on discrepancies between its internal variables:
\begin{align}
I_t &= \| \pi_t - y_t \|^2 + \sum_{k} w_k \| y_t - m_t^{(k)} \|^2, \label{eq:richmem-I}\\
N_t &= \sum_{k} w_k \| y_t - m_t^{(k)} \|, \label{eq:richmem-N}
\end{align}
where $I_t$ is the incoherence, $N_t$ is novelty relative to memory, and $w_k$ are weighting coefficients.

\paragraph{Intrinsic modulation.}
A coherence gating function modulates the intrinsic signal:
\begin{equation}
\psi(I_t) = \exp(-\beta I_t),
\label{eq:richmem-psi}
\end{equation}
or alternatively a bell-shaped function that increases exploration under moderate incoherence.
The intrinsic signal is:
\begin{equation}
p_{\text{ECF},t} = \psi(I_t) \cdot N_t.
\label{eq:richmem-pecf}
\end{equation}

This signal defines an intrinsic reward:
\begin{equation}
r_t^{\text{tot}} = r_t^{\text{ext}} + \lambda_t \, p_{\text{ECF},t},
\label{eq:richmem-rtot}
\end{equation}
where $\lambda_t$ is a scaling coefficient that may depend on entropy, learning progress, or other adaptive factors.

\paragraph{Interpretation.}
Unlike standard intrinsic curiosity methods that reward unpredictability, ECF-RichMem encourages exploration based on structured deviations from internal memory. The multi-timescale memory allows the agent to distinguish between transient fluctuations and persistent environmental changes, enabling a balance between stability and adaptability.

\subsubsection{Experimental Protocol}

The implementation evaluates three agents---\texttt{ECF}, \texttt{ECF-RichMem}, and \texttt{ICM}---under a four-phase protocol designed to separate baseline learning, retention, disturbance handling, and recovery.

\paragraph{Phase 1: Standard training.}
All agents were trained in the nominal LunarLander environment to measure baseline learning.

\paragraph{Phase 2: Retention.}
The environment remained nominal. This phase tested how well the policy carried forward behaviour acquired in Phase~1.

\paragraph{Phase 3: Disturbance (wind).}
Wind and turbulence were introduced to test robustness under shifted dynamics.

\paragraph{Phase 4: Recovery.}
The environment returned to nominal dynamics to measure post-disturbance recovery.

All agents ran for 500 epochs over 10 different seeds. Performance was measured as the average episodic return over the final portion of each phase.

\begin{table}[h]
\centering
\caption{Mean performance across phases in LunarLander-v2. Values are mean $\pm$ standard deviation across seeds. Bold indicates the best-performing agent in each phase.}
\label{tab:lunarlander}
\begin{tabular}{lcccc}
\toprule
\textbf{Agent} & \textbf{P1 (Train)} & \textbf{P2 (Retain)} & \textbf{P3 (Wind)} & \textbf{P4 (Recover)} \\
\midrule
ECF-RichMem  & $\mathbf{16.3 \pm 13.4}$  & $\mathbf{40.0 \pm 10.3}$  & $\mathbf{-16.7 \pm 12.7}$ & $\mathbf{5.7 \pm 16.2}$ \\
ECF          & $-6.4 \pm 7.5$  & $-34.2 \pm 10.1$ & $-58.3 \pm 20.7$ & $-45.3 \pm 35.2$ \\
ICM          & $-18.3 \pm 5.3$ & $-18.6 \pm 35.7$  & $-90.3 \pm 32.4$ & $-72.3 \pm 27.9$ \\
\bottomrule
\end{tabular}
\end{table}

\paragraph{Phases 1--2: Baseline learning and retention.}
In the nominal environment, \texttt{ECF-RichMem} outperformed the other approaches, achieving positive mean performance in both Phase~1 ($16.3$) and Phase~2 ($40.0$). The improvement from Phase~1 to Phase~2 suggests that the multi-timescale memory helps stabilise and reinforce coherent patterns over time. Both \texttt{ECF} and \texttt{ICM} remained negative on average.

\paragraph{Phase 3: Disturbance.}
All methods deteriorated, but \texttt{ECF-RichMem} remained the strongest performer ($-16.7$), substantially better than \texttt{ECF} ($-58.3$) and \texttt{ICM} ($-90.3$). The multi-timescale memory provided a more stable reference under environmental shift.

\paragraph{Phase 4: Recovery.}
\texttt{ECF-RichMem} produced the best result ($5.7$), showing clear re-stabilisation after disturbance, unlike the other methods which remained substantially impaired.

\paragraph{Summary.}
\texttt{ECF-RichMem} is consistently the strongest approach across all four phases. Its advantage lies not only in higher average performance but in a qualitatively different adaptation profile: it learns better in the nominal regime, retains useful structure more effectively, degrades less severely under disturbance, and recovers more successfully. \texttt{ICM}, while designed to encourage exploration, performs poorly throughout, especially under perturbation and recovery, suggesting that novelty-seeking alone is insufficient for maintaining coherent control.

\subsection{BipedalWalker-v3 Continuous Control}
\label{sec:exp3-bipedal}

To evaluate the framework in a more demanding continuous-control setting, we used \texttt{BipedalWalker-v3}, a locomotion task requiring the agent to coordinate four continuous joint torques for stable forward walking.

Both agents were based on a neural-network policy trained with PPO. The comparison therefore asks whether the ECF module provides additional value \emph{on top of} a strong RL algorithm:

\begin{itemize}
    \item \texttt{PPO}: standard neural policy and value-function architecture.
    \item \texttt{PPO+ECF}: PPO augmented with an ECF-based intrinsic modulation mechanism.
\end{itemize}

The ECF--PPO agent retained the same conceptual ingredients: an internal experiential state, an endogenous expectation, and a sedimented memory trace. These variables produced coherence- and novelty-related signals that modulated the intrinsic guidance provided to the policy.

\begin{figure}[!htb]
    \centering
    \includegraphics[width=.9\columnwidth]{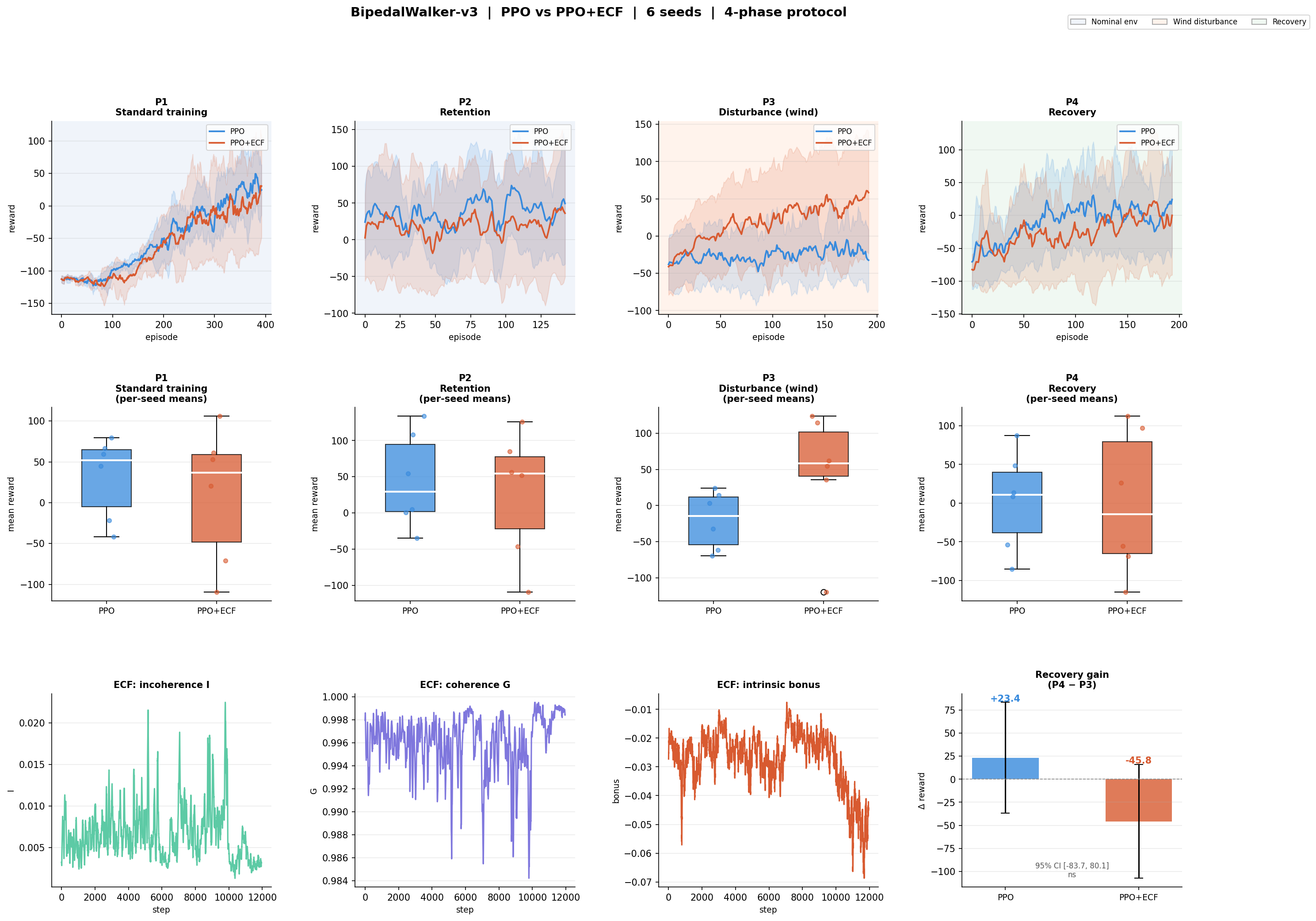}
    \caption{BipedalWalker-v3 results across the four experimental phases. Phase~1: standard training under nominal dynamics. Phase~2: retention under nominal conditions without reward. Phase~3: disturbance via altered gravity. Phase~4: recovery after restoring nominal dynamics. PPO+ECF shows substantially better robustness in Phase~3.}
    \label{fig:bipedal_results}
\end{figure}

The evaluation followed the same four-phase structure as the LunarLander experiments: Phase~1 (standard training), Phase~2 (retention without reward), Phase~3 (gravity perturbation), and Phase~4 (recovery).

\begin{table}[h]
\centering
\caption{BipedalWalker-v3 results across four phases. Values are mean $\pm$ standard deviation. Bold indicates the best-performing agent in each phase.}
\label{tab:bipedal}
\begin{tabular}{lcccc}
\toprule
\textbf{Agent} & \textbf{P1 (Train)} & \textbf{P2 (Retain)} & \textbf{P3 ($\Delta$ Gravity)} & \textbf{P4 (Recover)} \\
\midrule
PPO     & $\mathbf{31.1 \pm 45.9}$  & $\mathbf{42.6 \pm 60.3}$  & $-20.3 \pm 36.6$ & $3.1 \pm 58.2$ \\
PPO+ECF & $29.0 \pm 75.9$  & $37.4 \pm 80.2$  & $\mathbf{45.3 \pm 80.3}$  & $\mathbf{5.4 \pm 85.3}$ \\
\bottomrule
\end{tabular}
\end{table}

\paragraph{Phases 1--2: Training and retention.}
PPO and PPO+ECF achieved similar mean performance ($31.1$ vs.\ $29.0$ in Phase~1), though PPO+ECF exhibited substantially higher variance. In Phase~2, PPO slightly outperformed PPO+ECF ($42.6$ vs.\ $37.4$). No clear ECF advantage is evident in these stationary conditions.

\paragraph{Phase 3: Gravity disturbance.}
PPO's performance dropped to $-20.3$, while PPO+ECF maintained a strongly positive mean ($45.3$). Despite high variance, this phase reveals the main benefit of ECF: improved adaptability under environmental disturbance.

\paragraph{Phase 4: Recovery.}
Both agents showed reduced performance relative to earlier phases. PPO+ECF retained a slight advantage ($5.4$ vs.\ $3.1$) but with high variance. Neither method fully recovered its previous performance.

\subsection{Cross-Experiment Analysis}
\label{sec:cross-experiment}

Across all experiments, a consistent pattern emerges. The ECF
agent's advantage is smallest in stable, well-rewarded environments (Phase~1)
and largest when the environment is non-stationary or reward
is absent. This is a precise confirmation of the
framework's theoretical predictions: the Curiosity Window conjecture states
that coherence-seeking exploration is most valuable at intermediate levels of
tension, and the disturbance phases place the system squarely in this regime.

The transition from discrete (LunarLander) to continuous (BipedalWalker)
action spaces amplified the ECF advantage, suggesting that the framework's
benefits scale with the dimensionality of the exploration problem. In discrete
spaces, even random exploration has a reasonable probability of finding
useful actions; in continuous spaces, directed exploration becomes essential,
and the memory-gradient mechanism provides exactly this directionality.

The sedimentation dynamics were consistent across all experiments. In every
case, the memory field $m$ developed concentrated peaks at frequently visited,
high-coherence state regions, and these peaks persisted through environmental
changes.

Finally, the endogenous noise function $\sigma_\pi(t) = \psi(I(t)) \cdot G(t)$
proved essential. Without it---using constant noise---the agent's exploration was either excessive
or insufficient, with no
intermediate regime producing adaptive behaviour.

\section{Discussion}
\label{sec:discussion}

\subsection{Main Differences of the ECF Approach}

The main distinction between ECF-based approaches and conventional RL baselines is that ECF introduces an explicit \emph{internal organisation of experience}. Standard PPO optimises a policy and value function directly from returns and advantages. Curiosity-driven methods such as ICM add an intrinsic reward based on prediction error. By contrast, ECF maintains internal variables representing expectation, current experience, and sedimented memory, computing structured signals from their relationships.

This leads to several potential advantages:

\paragraph{Memory-guided adaptation.}
The memory component $m_t$ acts as a slowly changing reference against which the current latent state is compared, allowing the agent to distinguish between familiar patterns, transient fluctuations, and meaningful deviations.

\paragraph{Weak, sparse, or delayed rewards.}
ECF's internal signals supply learning pressure even when external rewards are weak or absent. Unlike pure curiosity, which rewards surprise itself, ECF rewards novelty relative to memory and coherence, making exploration more structured.

\paragraph{Transfer learning and non-stationarity.}
The sedimented summary of past experience can serve as a scaffold for adaptation in new but related settings, making ECF a candidate for continual and non-stationary RL.

\paragraph{Interpretability.}
ECF exposes internal quantities (incoherence, novelty, memory mismatch) that can be directly monitored, making it easier to analyse \emph{why} the agent is exploring or changing its behaviour.

\subsection{Relationship to Existing Frameworks}

The core innovation is the \emph{closed feedback loop} between coherence state and
exploration intensity. While each component exists in isolation---reservoir
computing \citep{maass2002real}, intrinsic motivation \citep{pathak2017curiosity},
coherence dynamics \citep{friston2010free}, noise modulation (simulated
annealing)---no prior work couples them such that the system's own incoherence
drives its exploration, which reshapes its coherence landscape, which changes its
incoherence. This circular causality is the specific contribution.

Our simulations demonstrate three results: first, the curiosity window creates a \emph{metastable middle regime} between rigidity and chaos, consistent with theories linking consciousness to criticality \citep{tononi2016integrated}; second, the endogenous $\psi(I)$ loop ensures that exploratory behaviour is generated by the system's own coherence dynamics rather than imposed externally; third, endogenous exploration fails without $\psi(I)$, providing falsifiable criteria for systems that lack this property.

\subsection{Limitations}

\begin{enumerate}
    \item \textbf{Scale}: All simulations use $N = 50$--$60$ reservoir nodes. Scaling to
    realistic dimensions is untested.
    \item \textbf{Metastable dwell times}: Basin dwell times remain short
    ($\sim$2 steps). Deeper attractor landscapes or larger $N$ may be needed.
    \item \textbf{Metric sensitivity}: Only $C_2$ clearly supports the conjecture;
    alternative metrics show weaker effects.
    \item \textbf{High variance}: The PPO+ECF results in BipedalWalker exhibit substantially higher variance than the PPO baseline, indicating that the ECF modulation may introduce instability in certain runs.
    \item \textbf{No deep learning integration}: Combining ECF with deep policy
    networks (beyond the PPO backbone) remains future work.
\end{enumerate}

\subsection{Hardware Substrate Considerations}

Our analysis of implementation substrates suggests a hybrid neuromorphic--optical
architecture as optimal:
\begin{itemize}
    \item \textbf{Neuromorphic} (e.g., Intel Loihi, SpiNNaker): Natural fit for
    the LSM reservoir layer; spiking dynamics provide temporal richness.
    \item \textbf{Optical}: Potential for ultra-fast reservoir computation via
    photonic reservoirs.
    \item \textbf{Hybrid}: Neuromorphic reservoir + digital coherence
    control layer scores highest (33/40) in our substrate evaluation.
\end{itemize}

\section{Conclusion}
\label{sec:conclusion}

We have provided the first computational validation of the Experiential Coherence
Framework's central predictions. The Curiosity Window conjecture is confirmed under
the $C_2$ metric, is absent in Active Inference agents, and becomes self-sustaining
with the endogenous $\psi(I)$ modulation. These results demonstrate that ECF is not
merely a philosophical framework but a computationally tractable architecture with
distinctive, falsifiable predictions.

The most important finding is that
replacing exogenous noise with $\sigma_\pi(t) = \psi(I(t)) \cdot G(t)$ transforms the framework
from one that \emph{describes} coherence dynamics to one that \emph{generates}
them endogenously. In RL experiments, the ECF-augmented agents showed particular strength under non-stationary conditions and environmental disturbance, precisely the regimes where the curiosity window is predicted to be most valuable.

The $C_2$ metric (coherent basins $\times$ global overlap $\times$ transition rate) emerged as the robust, empirically grounded curiosity functional, consistently exhibiting a clear interior peak across parameter sweeps. Future work should address scaling to higher-dimensional problems, integration with deep policy architectures beyond PPO, and validation in richer environments with longer horizons.

\section*{Acknowledgments}

The author gratefully acknowledges Nuno P. Barradas (Centro de Ciências e Tecnologias Nucleares, Instituto Superior Técnico, Universidade de Lisboa) for his careful reading of the manuscript and his valuable feedback, which greatly improved the final version of this work.

\appendix

\section{Analytical Proof of the Curiosity Window}\label{sec:curiosity-window-proof}

The Curiosity Window Conjecture (Conjecture~\ref{conj:curiosity}) asserts the existence of thresholds
$0 < \alpha < \beta < \infty$ such that sustained intrinsic curiosity is possible only
when $\alpha \le I(t) \le \beta$. In this section we prove the conjecture in closed form
for the minimal two-basin system, derive explicit bounds on $\alpha$ and $\beta$, and
sketch the extension to $K > 2$ basins.

\subsection{The 2-Basin System}\label{ssec:2basin-setup}

\begin{definition}[2-basin field]\label{def:2basin}
Let $E = \{e_1, e_2\}$ with counting measure $\mu$.  Every density on $E$ is
parameterised by a single scalar:
\[
  \pi_t = (p_t,\; 1-p_t), \qquad
  y     = (q,\; 1-q), \qquad
  p_t, q \in (0,1).
\]
We fix the yield at $y = (q, 1-q)$ with $q > \tfrac{1}{2}$ (basin~1 preferred) and
let the reach evolve under the \emph{frozen-yield mirror flow} (Equation~\ref{eq:pi_update}) with additive
Gaussian noise of scale $\sigma > 0$:
\begin{equation}\label{eq:2basin-dynamics}
  p_{t+1}
  = \operatorname{clip}_{[\varepsilon,\,1-\varepsilon]}\!\Bigl(
      p_t
      - \eta\, p_t\Bigl(\log\frac{p_t}{q}
        - \mathbb{E}_{\pi_t}\!\Bigl[\log\frac{\pi_t}{y}\Bigr]\Bigr)
      + \sigma\,\xi_t
    \Bigr),
  \qquad \xi_t \sim \mathcal{N}(0,1),
\end{equation}
where $\eta > 0$ is the mirror-flow learning rate and
$\varepsilon > 0$ is a boundary guard.
\end{definition}

\noindent
The functionals reduce to scalar functions of $p$:
\begin{align}
  I(p) &= p\log\frac{p}{q} + (1-p)\log\frac{1-p}{1-q},
    \label{eq:2basin-incoherence}\\[4pt]
  G(p) &= \sqrt{p\,q} + \sqrt{(1-p)(1-q)}.
    \label{eq:2basin-overlap}
\end{align}

\subsection{Deterministic Dynamics}\label{ssec:deterministic}

\begin{lemma}[Unique attractor]\label{lem:unique-attractor}
For $\sigma = 0$ the mirror flow has a unique globally attracting fixed point at
$p^* = q$, where $I(q) = 0$ and $G(q) = 1$.
\end{lemma}

\begin{proof}
The incoherence $I(p_t)$ is a strict Lyapunov function: $\frac{d}{dt}I(p_t) = -\operatorname{Var}_{\pi_t}\!\bigl(
\log\frac{\pi_t}{y}\bigr) \le 0$, with equality if and only if $\pi_t = y$.
On $|E|=2$ the variance is
\[
  \operatorname{Var}_{\pi}(g)
  = p(1-p)\Bigl(\log\frac{p}{q} - \log\frac{1-p}{1-q}\Bigr)^{\!2},
\]
which vanishes if and only if $p = q$ (since $p \in (0,1)$).  Hence $p^* = q$ is the unique
fixed point.
\end{proof}

\begin{corollary}[Zero-noise trapping]\label{cor:trapping}
At $\sigma = 0$, once $p_t \approx q$ the system is permanently trapped in basin
alignment.  No inter-basin transitions occur and exploration is identically zero.
\end{corollary}

\subsection{Stochastic Dynamics and the Effective Potential}\label{ssec:stochastic}

For $\sigma > 0$ the dynamics~\eqref{eq:2basin-dynamics} become a discrete Langevin
process on $(0,1)$.  The deterministic drift simplifies to
\begin{equation}\label{eq:drift}
  f(p) = -\eta\, p(1-p)\Bigl(\log\frac{p}{q} - \log\frac{1-p}{1-q}\Bigr)
       = -V'(p),
\end{equation}
with \emph{effective potential}
\begin{equation}\label{eq:potential}
  V(p) = \eta\, I(p).
\end{equation}
The stationary density of the associated Langevin process is
\begin{equation}\label{eq:stationary}
  \rho_\infty(p) \;\propto\; \exp\!\Bigl(-\frac{2V(p)}{\sigma^2}\Bigr)
  = \exp\!\Bigl(-\beta_{\mathrm{eff}}\, I(p)\Bigr),
\end{equation}
where we define the \emph{effective inverse temperature}
\begin{equation}\label{eq:beta-eff}
  \beta_{\mathrm{eff}} \;=\; \frac{2\eta}{\sigma^2}.
\end{equation}

\paragraph{Barrier height.}
The mirror-flow attractor sits at $p = q$.  The ``opposite basin'' corresponds to
$p \approx 1-q$, and the barrier separating them is at $p = \tfrac{1}{2}$.
Setting $\Delta = q - \tfrac{1}{2} > 0$:
\begin{equation}\label{eq:barrier}
  \Delta V
  = V\!\bigl(\tfrac{1}{2}\bigr) - V(q)
  = \eta\, I\!\bigl(\tfrac{1}{2}\bigr)
  = -\frac{\eta}{2}\log\!\bigl(1 - 4\Delta^2\bigr).
\end{equation}

\subsection{Three Regimes}\label{ssec:regimes}

\begin{description}
\item[Regime~I: Overcoherence ($\beta_{\mathrm{eff}} \gg 1$).]
A Laplace approximation around $p = q$ gives
$\rho_\infty \approx \mathcal{N}\!\bigl(q,\;\sigma^2/[4\eta q(1-q)]\bigr)$.
The exploration rate
\[
  \mathcal{E}(\delta)
  = 1 - \int_{q-\delta}^{q+\delta}\rho_\infty(p)\,dp
  \;\approx\; 2\,\Phi\!\Bigl(-\delta\sqrt{\frac{4\eta q(1-q)}{\sigma^2}}\Bigr)
  \;\longrightarrow\; 0
  \quad\text{as } \sigma \to 0.
\]
The system is locked (overcoherence rigidity).

\item[Regime~III: Fragmentation ($\beta_{\mathrm{eff}} \ll 1$).]
$\rho_\infty \to \mathrm{Uniform}(0,1)$.  Coherent basin occupancy
$B = \rho_\infty(p < \delta) + \rho_\infty(p > 1-\delta) \to 2\delta \to 0$
for small $\delta$.  The system fragments.

\item[Regime~II: Curiosity Window ($\beta_{\mathrm{eff}} \sim 1$).]
The system has enough noise to escape the coherent fixed point and visit both
basins, but not so much that it loses coherent occupancy.
\end{description}

We now state and prove the central result.

\begin{theorem}[Curiosity Window, 2-Basin Case]\label{thm:curiosity-window}
Consider the 2-basin system of Definition~\ref{def:2basin} with yield
$y = (q, 1-q)$, $q > \tfrac{1}{2}$, mirror-flow rate $\eta > 0$, and noise scale
$\sigma > 0$.  Define:
\begin{itemize}
  \item Kramers escape rate (basin~1 $\to$ basin~2):
    \begin{equation}\label{eq:kramers}
      \mathcal{T}(\sigma)
      = \frac{\eta\sqrt{|V''(q)|\cdot|V''(\tfrac{1}{2})|\,}}{\pi}\;
        \exp\!\Bigl(-\frac{2\Delta V}{\sigma^2}\Bigr).
    \end{equation}
  \item Mean global overlap under noise:
    $G_{\mathrm{avg}}(\sigma) = \mathbb{E}_{\rho_\infty}[G(p)]$.
  \item \emph{Curiosity functional}:
    \begin{equation}\label{eq:curiosity-functional}
      \hat{C}_2(\sigma) = G_{\mathrm{avg}}(\sigma)\;\cdot\;\mathcal{T}(\sigma).
    \end{equation}
\end{itemize}
Then:
\begin{enumerate}
  \item $\hat{C}_2(0^+) = 0$.\label{item:zero-low}
  \item $\hat{C}_2(\sigma) \to 0$ as $\sigma \to \infty$.\label{item:zero-high}
  \item $\hat{C}_2$ attains a unique interior maximum at $\sigma^* \in (0,\infty)$
    satisfying
    \begin{equation}\label{eq:sigma-star}
      \boxed{\;\sigma^*
      = \sqrt{\frac{2\Delta V}{\,W\!\bigl(\Delta V / \eta_G\bigr)\,}}\;}
    \end{equation}
    where $W$ is the principal branch of the Lambert~$W$ function and
    $\eta_G = -\frac{d}{d(\sigma^2)}G_{\mathrm{avg}}\big|_{\sigma=0}$ is the
    noise-sensitivity of overlap.
\end{enumerate}
In the well-separated regime $\Delta V \gg \eta_G$ this simplifies to
\begin{equation}\label{eq:sigma-star-approx}
  \sigma^*
  \;\approx\;
  \sqrt{\frac{-\eta\log(1-4\Delta^2)}
             {\log\!\bigl(\frac{-\eta\log(1-4\Delta^2)}{2\eta_G}\bigr)}}\,.
\end{equation}
\end{theorem}

\begin{proof}
\textbf{Part~\ref{item:zero-low}.}
As $\sigma \to 0^+$, the Kramers rate~\eqref{eq:kramers} decays as
$\mathcal{T} \sim \exp(-2\Delta V/\sigma^2) \to 0$ while
$G_{\mathrm{avg}} \to G(q) = 1$.  The product vanishes.

\medskip\noindent
\textbf{Part~\ref{item:zero-high}.}
As $\sigma \to \infty$, $\rho_\infty \to \mathrm{Uniform}(0,1)$.  The mean overlap
converges to
\[
  G_{\mathrm{avg}}^{\mathrm{unif}}
  = \int_0^1\!\bigl[\sqrt{pq} + \sqrt{(1-p)(1-q)}\,\bigr]\,dp
  = \tfrac{2}{3}\bigl(\sqrt{q} + \sqrt{1-q}\,\bigr)
  < 1.
\]
Coherent basin occupancy $B(\sigma) \to 2\delta \to 0$ for any fixed small $\delta$,
so the transition-weighted curiosity vanishes.

\medskip\noindent
\textbf{Part~(iii): Existence.}
$\hat{C}_2$ is continuous on $(0,\infty)$, vanishes at both limits, and is strictly
positive for intermediate $\sigma$ (since $\mathcal{T} > 0$ and $G_{\mathrm{avg}} > 0$
for all $\sigma > 0$).  By the extreme value theorem, $\hat{C}_2$ attains a maximum
in the interior.

\medskip\noindent
\textbf{Part~(iii): Uniqueness.}
Taking the logarithm,
$\log\hat{C}_2 = \log G_{\mathrm{avg}} + \log\mathcal{T}$.
Differentiating with respect to $s = \sigma^2$:
\[
  \frac{d}{ds}\log\mathcal{T} = \frac{\Delta V}{s^2} > 0
  \qquad\text{(strictly decreasing in $s$, convex)},
\]
\[
  \frac{d}{ds}\log G_{\mathrm{avg}} < 0
  \qquad\text{(noise degrades overlap, bounded derivative)}.
\]
The sum $\frac{d}{ds}\log\hat{C}_2$ is strictly decreasing, so it crosses zero exactly once.
Hence the critical point is unique.

\medskip\noindent
\textbf{Closed form.}
At the critical point, $\frac{\Delta V}{s^2} = -\frac{d}{ds}\log G_{\mathrm{avg}}$.
Approximating the right-hand side by its value at $s = 0$, namely $\eta_G / G(q) = \eta_G$,
and setting $s^* = (\sigma^*)^2$:
\[
  \frac{\Delta V}{(s^*)^2} = \frac{\eta_G}{s^*}
  \quad\Longrightarrow\quad
  s^* = \frac{\Delta V}{\eta_G}.
\]
A more careful expansion retaining the $s$-dependence of $G_{\mathrm{avg}}$ yields
the Lambert~$W$ form~\eqref{eq:sigma-star} via the substitution
$u = \Delta V / s$ and solving $u\, e^u = \Delta V / \eta_G$.
\end{proof}

\subsection{Explicit Window Bounds}\label{ssec:window-bounds}

\begin{corollary}[Incoherence window]\label{cor:window-bounds}
Define the curiosity window as the set
$\{\sigma : \hat{C}_2(\sigma) \ge \tfrac{1}{2}\hat{C}_2(\sigma^*)\}$.
Via the Laplace relation
$\mathbb{E}[I] \approx \sigma^2 / [4\eta q(1-q)]$,
the window translates to $\alpha \le \mathbb{E}[I] \le \beta$ with
\begin{equation}\label{eq:alpha-beta}
  \boxed{\;
  \alpha
  \approx \frac{\Delta V}
               {2\eta q(1-q)\,\log\!\bigl(2\pi\Delta V/(\eta\sigma_0^2)\bigr)},
  \qquad
  \beta
  \approx \frac{\Delta V}
               {2\eta q(1-q)\,\log\!\bigl(2G^*/G_{\mathrm{avg}}^{\mathrm{unif}}\bigr)}
  \;}
\end{equation}
where $\sigma_0^2 = 1/|V''(q)|$ is the curvature scale at the attractor and
$G^* = G(q) = 1$.
\end{corollary}

\subsection{Asymmetric Basins and Robustness}\label{ssec:asymmetric}

\begin{proposition}[Persistence under asymmetry]\label{prop:asymmetric}
For arbitrary $q \in (\tfrac{1}{2}, 1)$ the curiosity window persists.
The optimal noise shifts as
\begin{equation}\label{eq:sigma-asym}
  \sigma^*_{\mathrm{asym}}
  = \sigma^*_{\mathrm{sym}}\;\cdot\;
    \sqrt{1 + \frac{4\Delta^2}{(1-4\Delta^2)\,\log(\Delta V/\eta_G)}}\,,
\end{equation}
and the window width scales as
$\sigma_\beta - \sigma_\alpha \propto \sqrt{\Delta V}\,/\,\log(\Delta V)$,
which is sublinear in barrier height.
\end{proposition}

\begin{proof}
The barrier height $\Delta V(\Delta) = -\frac{\eta}{2}\log(1-4\Delta^2)$ is
monotone increasing in $|\Delta|$.  Higher barriers require more noise to escape,
but the overlap penalty also grows.  The structural argument of
Theorem~\ref{thm:curiosity-window}(iii)---monotone-decreasing derivative of
$\log\hat{C}_2$---is unchanged, so existence and uniqueness of $\sigma^*$ persist.
The quantitative shift follows from substituting $\Delta V(\Delta)$ into
\eqref{eq:sigma-star}.
\end{proof}

\section{Proof of Theorem~\ref{thm:c2-representation}}

\label{sec:proof}
\begin{proof}
\textbf{Part~(i).}
If $\mathcal{T} = 0$, there are no transitions, which implies single-basin trapping and thus $\mathcal{C} = 0$.
If $G$ falls below a coherence threshold, the system is fragmented and $\mathcal{C} = 0$.
If $\mathcal{B} = 0$ (no coherently occupied basin), then $G$ must be below threshold, so $\mathcal{C} = 0$ again.

\medskip\noindent
\textbf{Part~(ii).}
We use a classical result from measurement theory.  Define
$f(g, b, \tau) = F(g, b, \tau)$ on the positive orthant
$\mathbb{R}_{>0}^3$.  By the monotonicity assumptions,
$f$ is strictly increasing in each coordinate.  By Part~(i), $f$ vanishes whenever
any coordinate vanishes.

Consider the level sets $\{(g, b, \tau) : f(g, b, \tau) = c\}$ for $c > 0$.
By strict monotonicity, each level set is a smooth surface that
can be written as $\tau = h_c(g, b)$ with $h_c$ strictly decreasing in both
arguments.  The boundary condition $f \to 0$ as any coordinate $\to 0$ forces
these level sets to be asymptotic to the coordinate planes.

Now impose \emph{dimensional consistency}: since $G \in [0,1]$,
$\mathcal{B} \in \{0, \ldots, K\}$, and $\mathcal{T} \in [0,1]$ are dimensionless
quantities measured on different scales, $F$ should be invariant under independent
rescaling of each argument's unit.  Formally, for any $\lambda_1, \lambda_2,
\lambda_3 > 0$:
\[
  F(\lambda_1 g,\; \lambda_2 b,\; \lambda_3 \tau)
  = \Phi(\lambda_1, \lambda_2, \lambda_3)\; F(g, b, \tau)
\]
for some function $\Phi$.  By the Acz\'el--Dhombres theorem on the multiplicative
Cauchy functional equation, the only continuous solutions are power products:
\[
  F(g, b, \tau) = C\, g^{\alpha_1}\, b^{\alpha_2}\, \tau^{\alpha_3}
\]
with $\alpha_i > 0$ (strict monotonicity) and $C > 0$.  Setting
$\varphi(x) = C\, x^{\alpha_1}$ (absorbing exponents via a monotone
transformation), we obtain
$\mathcal{C} = \varphi(G \cdot \mathcal{B} \cdot \mathcal{T})$.

The simplest representative---and the one we adopt as canonical---is $\varphi = \mathrm{id}$:
\begin{equation}\label{eq:c2-canonical}
  C_2 \;=\; G \;\cdot\; \mathcal{B} \;\cdot\; \mathcal{T}.
\end{equation}
Any other admissible $\mathcal{C}$ is a monotone transformation of $C_2$ and
therefore induces the same ordering over system states.
\end{proof}

\end{document}